%% file: paper.tex
\documentclass[]{style}

\input{preamble}

\usepackage[utf8]{inputenc} 
\usepackage[T1]{fontenc}    
\usepackage{hyperref}       
\usepackage{url}            
\usepackage{booktabs}       
\usepackage{amsfonts}       
\usepackage{nicefrac}       
\usepackage{microtype}      
\usepackage{amsfonts}
\usepackage{amsmath}
\usepackage{graphicx}
\usepackage{url}
\usepackage{xspace}
\usepackage{amssymb}
\usepackage{amsmath}
\usepackage{pifont}
\usepackage{algorithm}
\usepackage{algorithmicx}
\usepackage{algpseudocode}
\usepackage{wrapfig}
\usepackage{booktabs}
\usepackage{multirow}
\usepackage[table]{xcolor}

\usepackage{amsmath}
\usepackage{soul}
\usepackage{tabularx}
\usepackage{tikz}
\usepackage{lipsum}
\usepackage{enumitem}
\usepackage[most]{tcolorbox}

\definecolor{mypink}{HTML}{D86ECC}
\definecolor{gitred}{RGB}{255,230,230}   
\definecolor{gitgreen}{RGB}{230,255,230} 
\definecolor{gitredtext}{RGB}{160,0,0}   
\definecolor{gitgreentext}{RGB}{0,100,0} 

\newcommand{\diffline}[4]{%
  \makebox[0pt][l]{%
    {\color{#1}\rule[-1.8\dp\strutbox]{\dimexpr\linewidth-1.1em\relax}{\dimexpr1.8\ht\strutbox+1.8\dp\strutbox\relax}}%
  }%
  \hspace{#3}\makebox[\dimexpr\linewidth-#3-2.2em\relax][l]{\strut#4}%
  \llap{\textbf{\texttt{#2}}\hspace{0.2em}}%
}
\newcommand{\gitrem}[2][0pt]{\diffline{gitred}{-}{#1}{\color{gitredtext}#2}}
\newcommand{\gitadd}[2][0pt]{\diffline{gitgreen}{+}{#1}{\color{gitgreentext}#2}}

\hypersetup{
    citecolor=YellowGreen,     
}

\definecolor{catgray}{gray}{0.92}

\algrenewcommand\algorithmicindent{1.2em}

\tcbset{
    calloutbox/.style={
        colback=blue!5!white,
        colframe=blue!60!black,
        coltitle=white,
        fonttitle=\bfseries,
        boxrule=0.8pt,
        arc=3pt,
        left=8pt,
        right=8pt,
        top=6pt,
        bottom=6pt,
        breakable,
        enhanced,
    }
}
 
\newcommand{\timeline}[1]{\textbf{\texttt{#1}}}

\definecolor{green}{HTML}{009000}
\definecolor{red}{HTML}{ea4335}

\makeatletter
\DeclareRobustCommand\onedot{\futurelet\@let@token\@onedot}
\def\@onedot{\ifx\@let@token.\else.\null\fi\xspace}

\def\eg{\emph{e.g}\onedot} 
\def\ie{\emph{i.e}\onedot} 
 
\def\etc{\emph{etc}\onedot}

\makeatother

\newcommand{\framework}{Delta Forcing\xspace}

\title{\framework: Trust Region Steering for Interactive Autoregressive Video Generation}
\author[1,*,\ddagger]{Yuheng Wu}
\author[3]{Xiangbo Gao}
\author[1]{Tianhao Chen}
\author[4]{Xinghao Chen}
\author[2]{Qin Ying}
\author[2,3,\ddagger]{Zhengzhong Tu}
\author[1]{Dongman Lee}

\affiliation[1]{KAIST}
\affiliation[2]{Visko Platform}
\affiliation[3]{Texas A\&M University}
\affiliation[4]{University of Washington}

\contribution[*]{Work done when intern at Visko}
\contribution[\ddagger]{Project Lead}

\abstract{
Interactive streaming video generation requires a causal model to react to evolving event conditions while preserving the visual trajectory it has already established.
A fundamental challenge lies in balancing \emph{reactivity} and \emph{stability}: models must respond promptly to new events while maintaining temporal coherence over long horizons. 
Existing approaches typically distill high-quality bidirectional diffusion models into autoregressive generators and further adapt them via streaming long-tuning; however, they often suffer from identity, layout, and scene drift when event switches. 
We identify the cause of this failure as \emph{conditional bias}, where the (frozen) teacher can produce denoising guidance that is semantically valid for the new event but incompatible with the student's realized history, thereby causing distribution matching to propagate prompt-consistent yet trajectory-inconsistent modes. 
We propose \framework, a transition-calibrated steering objective for a distribution-matching distillation.
For each generated chunk, Delta Forcing compares the feature-space delta of the teacher’s denoised estimate with the delta of the generator’s history-grounded rollout, retaining full teacher-driven distribution matching when the transitions agree and shifting toward trajectory-continuity regularization when they diverge. This adaptive supervision restores the balance between reactivity and stability, enabling coherent long-term generation under dynamic conditions. Extensive experiments demonstrate that \framework achieves the best subject/background consistency (96.60/94.63), the highest VideoAlign score (7.55), and the best user-study average rank (1.96), demonstrating improved long-horizon stability while maintaining responsiveness to evolving events.
}

\project{\url{https://delta-forcing-website.vercel.app/}}
\date{\today}
\contact{Yuheng Wu (\href{mailto:yuhengwu@kaist.ac.kr}{yuhengwu@kaist.ac.kr}), Zhengzhong Tu (\href{mailto:tzz@tamu.edu}{tzz@tamu.edu})}

\begin{document}
\thispagestyle{firstheader}
\maketitle
\pagestyle{plain}

\input{sec/1_intro}
\input{sec/2_related_work}

\input{sec/3_motivation}
\input{sec/4_method}

\input{sec/5_experiments}

\input{sec/6_discussion}

\input{sec/7_conclusion}

\newpage
{
\small
\bibliographystyle{IEEEtran}
\bibliography{main}
}

\clearpage
\newpage
\beginappendix
\input{sec/appendix}

\end{document}

%% file: preamble.tex
\definecolor{lightblue}{rgb}{0.0, 0.6, 1.0}
\definecolor{darkgreen}{rgb}{0.0, 0.6, 0.0}
\definecolor{lightgreen}{rgb}{0.1, 0.8, 0.1}
\definecolor{lightred}{rgb}{0.7, 0.7, 0.7}
\definecolor{lightgreen}{rgb}{0, 0, 0}
\definecolor{lightlightgray}{rgb}{0.8, 0.8, 0.8}

\usepackage{titlesec}
\titlespacing*{\paragraph}
{0pt}{0em}{0.2em}  

\makeatletter
\renewcommand\paragraph{\@startsection{paragraph}{4}{\z@}%
  {3pt}
  {-0.5em}
  {\normalfont\normalsize\bfseries}} 
\makeatother

\usepackage[utf8]{inputenc} 
\usepackage[T1]{fontenc}    
\usepackage{hyperref}       
\usepackage{url}            
\usepackage{booktabs}       
\usepackage{amsfonts}       
\usepackage{nicefrac}       
\usepackage{microtype}      
\usepackage{amsfonts}
\usepackage{amsmath}
\usepackage{graphicx}
\usepackage{url}
\usepackage[dvipsnames,table,xcdraw]{xcolor}
\usepackage{xspace}
\usepackage{amssymb}
\usepackage{amsmath}
\usepackage{pifont}
\usepackage{algorithm}
\usepackage{algorithmicx}
\usepackage{algpseudocode}
\usepackage{wrapfig}
\usepackage{booktabs}
\usepackage{multirow}
\usepackage[table]{xcolor}
\usepackage[numbers]{natbib}

\usepackage{amsmath}
\usepackage{soul}
\usepackage{tabularx}
\usepackage{tikz}
\usepackage{lipsum}
\usepackage{enumitem}
\usepackage[most]{tcolorbox}

\usepackage{booktabs}

\usepackage{url}
\definecolor{blue}{rgb}{0.21,0.49,0.74}
\definecolor{red}{rgb}{0.8, 0.2, 0.2}
\definecolor{green}{rgb}{0, 0.5, 0}
\definecolor{yellow}{RGB}{218, 160, 109}
\definecolor{gray}{RGB}{155, 155, 155}

\crefname{section}{Sec.}{Secs.}
\Crefname{section}{Section}{Sections}
\Crefname{table}{Table}{Tables}
\crefname{table}{Tab.}{Tabs.}
\crefname{figure}{Fig.}{Figs.}
\Crefname{figure}{Figure}{Figures}
\crefname{appendix}{App.}{Apps.}
\Crefname{appendix}{Appendix}{Appendices}

%% file: sec/1_intro.tex
\section{Introduction}
\label{sec:intro}
Interactive streaming video generation is emerging as a fundamental capability for applications such as real-time content creation and world modeling, where visual content must be continuously generated under dynamically evolving event conditions (\eg, prompts, actions, or control signals). Unlike offline generic video generation~\cite{wan2025wan,wu2025hunyuanvideo,hacohen2024ltx,seedance2026seedance,team2025kling,brooks2024video,lin2024open,bao2024vidu}, this setting introduces a fundamental tension between \emph{reactivity} and \emph{stability}: the model must respond instantaneously to new events while simultaneously preserving temporal consistency across events.

To make such systems practical, existing approaches typically adopt a two-stage pipeline. \textit{In stage 1}, high-quality but computationally expensive bidirectional models are distilled into efficient few-step autoregressive diffusion transformer (AR-DiT) generators via forcing-based approaches, such as Diffusion Forcing~\cite{chenDiffusionForcingNexttoken2024}, Self-Forcing~\cite{huangSelfForcingBridging2025,yinSlowBidirectionalFast2025}, and Causal Forcing~\cite{zhuCausalForcingAutoregressive2026} which commonly trained with distribution matching objectives~\cite{yinOnestepDiffusionDistribution2024,yin2024improved}. \textit{In stage 2}, methods such as LongLive~\cite{yangLongLiveRealtimeInteractive2025} and MemFlow~\cite{jiMemFlowFlowingAdaptive2025} further apply \emph{streaming long tuning}, fine-tuning such generators on long sequences with evolving event conditions to enable runtime interactivity.

Despite these advances, a persistent failure mode remains in current pipelines. The generated content, while still semantically aligned with the updated event conditions, gradually drifts in scene layout, object appearance and \etc. This persists even in memory-aware generators (\eg, MemFlow~\cite{jiMemFlowFlowingAdaptive2025}) that are explicitly designed to preserve long-term consistency. Previous studies~\cite{yangLongLiveRealtimeInteractive2025,jiMemFlowFlowingAdaptive2025,huangSelfForcingBridging2025,zhuCausalForcingAutoregressive2026,cui2025self} often attribute such degradation to the accumulation of autoregressive generation errors. In contrast, we identify a more fundamental cause: a systematic bias introduced across events by the teacher model during stage 2 training. \textbf{Existing approaches implicitly assume that the teacher provides reliable supervision under evolving event conditions, and thus enforce alignment with the teacher distribution at each event equally}. However, during stage 2 training, the teacher itself may accumulate errors and exhibit biased behavior, which in turn propagates to the student and induces drift. Empirically, when a new event is introduced (\eg, ``...grandfather enters...''$\rightarrow$``...talk back...''), the teacher may shift toward a condition-dependent distribution that faithfully reflects the updated semantics but fails to account for the previously generated trajectory (\eg, ``dark blue clothes''$\rightarrow$``white clothes'') as indicated by \textcolor{mypink}{\ding{202}}. Consequently, when such guidance is directly enforced via DMD~\cite{yinOnestepDiffusionDistribution2024,yin2024improved} (as indicated by \textcolor{red}{\ding{203}}), the model is repeatedly driven toward condition-consistent yet trajectory-inconsistent states. We term this phenomenon \textbf{\emph{conditional bias}}, where condition-driven shifts in the teacher distribution, coupled with inherent data priors, systematically steer generation toward trajectory-inconsistent modes.

\begin{figure}
    \centering
    \includegraphics[width=0.9\linewidth]{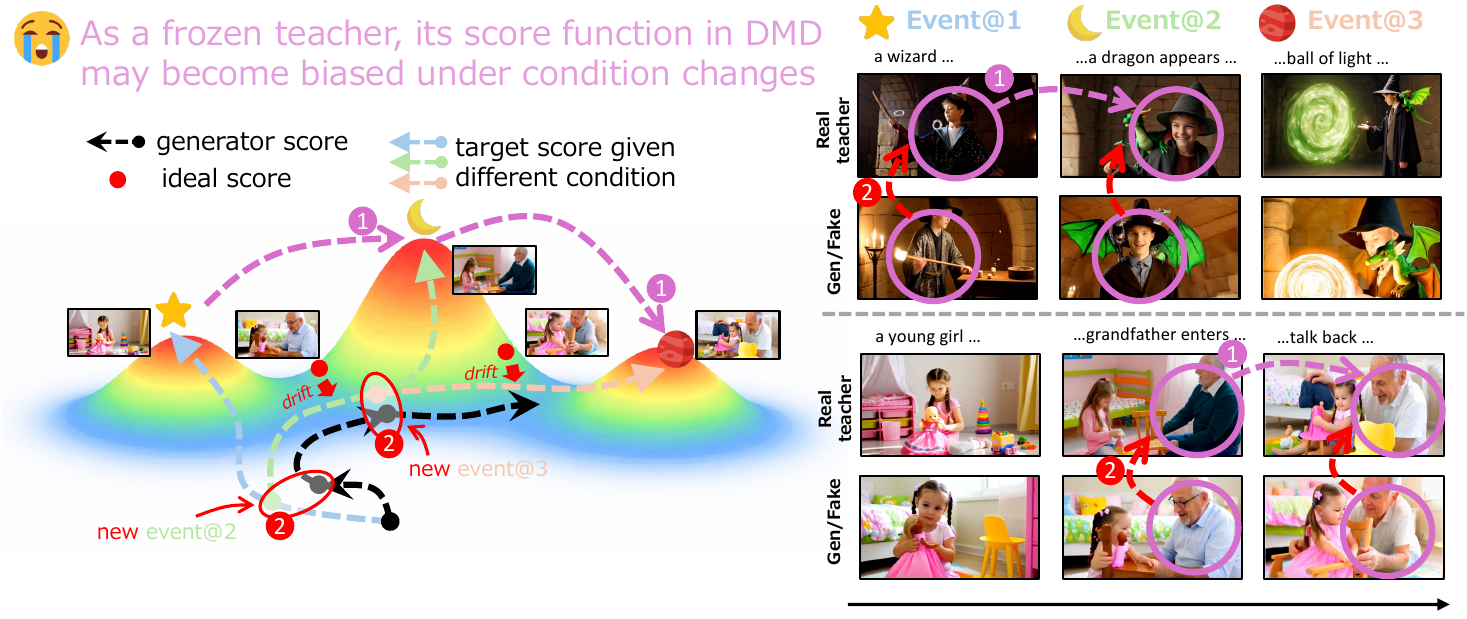}
    \caption{\textbf{Left:} Under evolving events, the frozen teacher, biased toward certain patterns, remains condition-aware but trajectory-agnostic, inducing \emph{conditional bias} that deviates from the historical trajectory. \textbf{Right:} Decoding both the real teacher model (\ie, Wan2.1-14B-T2V~\cite{wan2025wan}) and generator (MemFlow~\cite{jiMemFlowFlowingAdaptive2025}) shows that the generator’s drift closely follows these teacher-induced shifts.}
    \label{fig:intro}
\end{figure}

In this work, we propose \textbf{\framework}, a framework that restores the balance between \emph{reactivity} and \emph{stability} under conditional bias. Instead of directly matching the teacher's potentially biased distribution, we shift to \emph{reliability-aware transition modeling} by examining the transition it induces between consecutive latent states. Inspired by trust-region policy optimization~\cite{schulman2015trust}, where updates are constrained to a region in which the approximation remains reliable, we treat each teacher-induced transition as a candidate update and assess whether it falls within a trajectory-consistent trust region. Concretely, we introduce a delta-based mechanism that gauges the semantic shift of the teacher-induced transition relative to the generator's history-grounded trajectory, and uses this signal to modulate supervision online: transitions within the trust region are reinforced, while those that deviate are suppressed. 
This adaptive scheme effectively filters out bias-induced guidance, enabling the model to maintain coherent long-term evolution while remaining responsive to new conditions.

In summary, our contributions are threefold. \textit{First}, we identify \emph{conditional bias} in interactive multi-event streaming video generation, a failure mode in which event-driven condition updates steer the teacher toward trajectory-inconsistent modes, corrupting student supervision. \textit{Second}, inspired by trust-region policy optimization, we propose \framework, a reliability-aware framework that introduces a delta-based mechanism to modulate supervision online, reinforcing trajectory-consistent guidance while suppressing bias-induced deviations. \textit{Third}, extensive experiments demonstrate that \framework substantially improves temporal stability while preserving event reactivity in long-horizon multi-event scenarios.

%% file: sec/2_related_work.tex
\section{Related Work}
\label{sec:relaetdwork}

\subsection{Stability in Real-time Streaming Video Generation}

\begin{wrapfigure}{r}{0.5\linewidth}
    \vspace{-1em}
    \begin{center}
    \includegraphics[width=\linewidth]{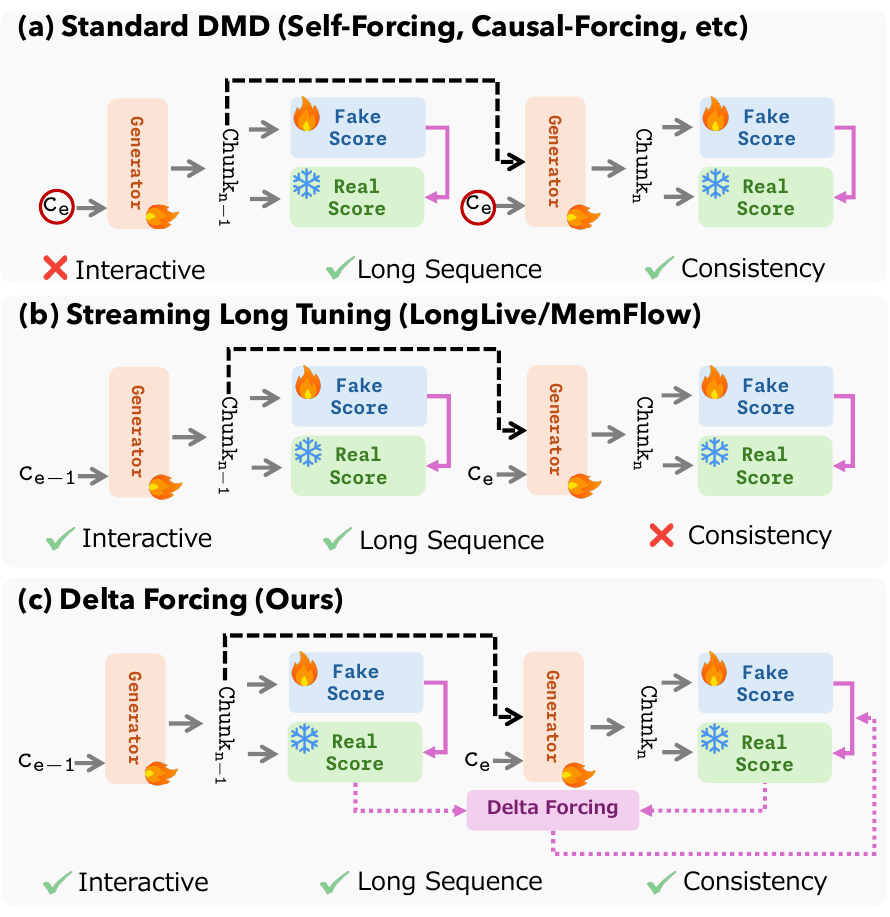}
    \end{center}
    \vspace{-0.5em}
    \caption{(a) Standard DMD fails to handle condition changes. (b) Streaming Long Tuning improves interactivity but still suffers from biased guidance, and (c) our method enforces transition consistency to mitigate conditional bias and preserve temporal coherence.}
    \label{fig:relatedwork}
    \vspace{-1.5em}
\end{wrapfigure}
Real-time streaming video generation requires producing temporally coherent content over long horizons under causal constraints, a setting fundamentally more demanding than offline generation. A promising paradigm distills a strong pretrained bidirectional DiT into a causal, few-step generator capable of efficient AR synthesis~\cite{huang2026live,chenDiffusionForcingNexttoken2024,huangSelfForcingBridging2025,cui2025self,chen2026context,zhuCausalForcingAutoregressive2026,liuRollingForcingAutoregressive2025,luRewardForcingEfficient2025,yang2026anchor,chen2026grounded,liu2026streamingautoregressivevideogeneration,zou2026hiar}. The central challenge in this paradigm is \emph{temporal stability}: errors accumulate over rollouts and manifest as content drift and structural inconsistency, a gap that widens with generation length. A line of work addresses this by narrowing the discrepancy between training and inference. Diffusion Forcing~\cite{chenDiffusionForcingNexttoken2024} trains a causal model with independent per-token noise levels, unifying clean past and noisy future frames to enable stable rollout beyond the training horizon. Self-Forcing~\cite{huangSelfForcingBridging2025} closes the train-test gap at its source by conditioning each frame on self-generated rollouts during training. Context Forcing~\cite{chen2026context} reveals a student--teacher context mismatch in streaming tuning and trains the AR student with a long-context teacher. Causal Forcing~\cite{zhuCausalForcingAutoregressive2026} identifies that distilling an AR student from a bidirectional teacher violates frame-level ODE injectivity and addresses this by initializing from an AR teacher, substantially recovering generation quality. Rolling Forcing~\cite{liuRollingForcingAutoregressive2025} jointly denoises a rolling window of frames with progressively increasing noise levels, anchoring global consistency via an attention sink over initial-frame KV states to sustain stability across multi-minute streams. Reward Forcing~\cite{luRewardForcingEfficient2025} targets the complementary issue that distilled models suffer diminished motion dynamics, introducing a rewarded distribution matching objective that biases training toward high-reward, high-motion samples. As illustrated in Fig.~\ref{fig:relatedwork}(a), despite their differences, these methods share a common assumption: the student is aligned with teacher guidance under a \emph{fixed} condition. This assumption breaks down in interactive settings where conditions evolve dynamically.

\subsection{Reactivity in Interactive Streaming Video Generation}
A complementary line of work extends AR video generation to interactive settings, where conditions evolve dynamically and the model must adapt to each new event without sacrificing temporal consistency. Early efforts approach interactivity through architectural and scheduling design. SkyReels-V2~\cite{chen2504skyreels} provides a diffusion-forcing substrate for long rollouts but assumes a fixed, pre-specified condition and does not address mid-rollout event changes. MAGI-1~\cite{teng2025magi} scales AR generation via chunk-wise prediction and exposes chunk-wise prompting as a controllability hook, but each prompt switch requires per-transition tuning of the KV attention range rather than being learned from streaming supervision. A more principled approach targets reactivity through \emph{streaming long tuning}, as illustrated in Fig.~\ref{fig:relatedwork}(b): the generator is fine-tuned on long self-rollouts with evolving conditions, directly aligning training with interactive inference. LongLive~\cite{yangLongLiveRealtimeInteractive2025} augments this with a KV-recache mechanism that updates stale context while preserving visual continuity at each event boundary. MemFlow~\cite{jiMemFlowFlowingAdaptive2025} replaces fixed memory strategies with prompt-guided dynamic retrieval of relevant historical context, improving narrative coherence across scene switches. Despite these advances, training on evolving conditions does not resolve the underlying supervision problem: at each event boundary, the teacher is steered toward a new condition-dependent distribution that may contradict the established trajectory, providing biased guidance that propagates through subsequent generation.

%% file: sec/3_motivation.tex
\section{Motivation Study}
\label{sec:motivation}

In this paper, we study interactive \emph{multi-event} streaming video generation, where a streaming generated video must adapt to dynamically evolving event conditions in real time. Unlike multi-shot generation, which transitions between events via hard cuts, our setting demands smooth, uncut evolution across events. We therefore begin with a key question: \emph{when drift emerges, where does the corrupting signal actually come from?}

\noindent\textbf{Preliminaries: enabling interactivity in streaming video generation.}
We follow LongLive~\cite{yangLongLiveRealtimeInteractive2025}, which adopts a two-stage recipe. In the \emph{first stage}, a causal, chunk-level autoregressive student $G_\theta$ with parameters $\theta$ is forced from a bidirectional diffusion teacher~\cite{huangSelfForcingBridging2025,zhuCausalForcingAutoregressive2026} via DMD~\cite{yinOnestepDiffusionDistribution2024,yin2024improved}, which adapts the teacher's bidirectional attention into a causal AR generator while compressing multiple denoising steps into a few. DMD aligns the student's induced distribution $p_{\mathrm{fake}}$ with the teacher's $p^\star$ by minimizing the time-averaged reverse KL, whose gradient takes the score-difference form
\begin{equation}
    \nabla_\theta \mathcal{L}_{\mathrm{DMD}}
    = -\mathbb{E}\!\left[
    \big(\mathbf{s}^\star(\mathbf{x}, t) - \mathbf{s}_{\mathrm{fake}}(\mathbf{x}, t)\big)
    \frac{\partial G_\theta}{\partial \theta}
    \right],
\end{equation}
where $\mathbf{x}$ is a noisy sample obtained by perturbing $G_\theta(\mathbf{z})$ at noise level $t$ (with $\mathbf{z}$ a Gaussian latent), and $\mathbf{s}^\star$, $\mathbf{s}_{\mathrm{fake}}$ are the scores $\nabla_\mathbf{x}\log p^\star_t$ and $\nabla_\mathbf{x}\log p_{\mathrm{fake},t}$ estimated by the frozen teacher and a learnable fake critic, respectively. This stage yields a low-latency AR student, but it is trained on short clips with a single static condition and offers no mechanism for handling event transitions.

In the \emph{second stage}, to enable multi-event interaction in training itself, LongLive adopts \emph{streaming long tuning}, which extends rollouts to long sequences containing condition switches. Let $e \in \{1, 2, \dots\}$ index events along a trajectory, $\mathbf{c}_e$ the condition (\eg, prompt) at event $e$, and $\mathbf{h}_{e-1}$ the cached history accumulated before event $e$. At each event, the student rolls out a new chunk on top of its \emph{own} history under the updated condition, and DMD supervision is applied locally:
\begin{equation}
    \mathcal{L}_{\mathrm{stream}}
    = \mathbb{E}_{e}\!\left[
    \mathrm{KL}\!\left(
    p_{\mathrm{fake}}(\mathbf{x} \mid \mathbf{h}_{e-1}, \mathbf{c}_e)
    \,\big\|\,
    p^\star(\mathbf{x} \mid \mathbf{c}_e)
    \right)
    \right],
\end{equation}
This rolling schedule exposes the student to its own self-generated history across events while bounding each gradient step to a single chunk.

\noindent\textbf{Observation: student drift is inherited from the teacher.} To identify the source of drift, we decode both the teacher's predicted outputs and the student's generations into pixel space and compare their trajectories. As illustrated in Fig.~\ref{fig:intro}(right), although DMD aligns distributions rather than pixels, the student tracks the teacher at the semantic level: whenever the teacher's output shifts to a new semantic mode in response to an updated $\mathbf{c}_e$ (\eg, a change in object appearance, spatial layout, or scene composition), the student's subsequent generations drift in the same direction. We further visualize this inconsistency and the accompanying semantic jumps in the latent feature trajectories of current SOTA models~\cite{yangLongLiveRealtimeInteractive2025,jiMemFlowFlowingAdaptive2025,luRewardForcingEfficient2025} in Appendix~\ref{apx:motivation}. Drift is thus not random error accumulation but is \emph{inherited} from the teacher's supervision signal.

\noindent\textbf{Diagnosis: trajectory-agnostic supervision under history-aware dynamics.} The ideal generative target depends on both $\mathbf{h}_{e-1}$ and $\mathbf{c}_e$:
\begin{equation}
    p^\star(\mathbf{x} \mid \mathbf{h}_{e-1}, \mathbf{c}_e),
    \qquad
    \mathbf{s}^\star(\mathbf{x}, t \mid \mathbf{h}_{e-1}, \mathbf{c}_e)
    =
    \nabla_{\mathbf{x}} \log p_t^\star(\mathbf{x} \mid \mathbf{h}_{e-1}, \mathbf{c}_e),
\end{equation}
where $\mathbf{s}^\star(\mathbf{x}, t \mid \mathbf{h}_{e-1}, \mathbf{c}_e)$ is the history-aware teacher score. Under this target, event transitions act as \emph{local deformations} of the trajectory, with state and update direction consistent with the incoming history. Pretrained video teachers (\eg, Wan~\cite{wan2025wan}, Hunyuan~\cite{wu2025hunyuanvideo}), however, are trained on short clips with a single static condition and provide no mechanism for cross-event conditional changes. The teacher therefore may drift across events, which is transferred through distillation into the student.

\noindent\textbf{Mechanism: conditional bias.} Because the teacher cannot distinguish among incoming histories, the score the student effectively matches is the history-marginalized
\begin{equation}
    \bar{\mathbf{s}}^\star(\mathbf{x}, t \mid \mathbf{c}_e)
    =
    \mathbb{E}_{\mathbf{h}_{e-1} \sim p(\cdot \mid \mathbf{c}_e)}\!\left[
    \mathbf{s}^\star(\mathbf{x}, t \mid \mathbf{h}_{e-1}, \mathbf{c}_e)
    \right].
\end{equation}
Trajectory-dependent information is averaged out, so the student is no longer constrained to follow its incoming trajectory and is instead pulled toward condition-consistent modes that may correspond to entirely different ones (Fig.~\ref{fig:intro}). We refer to this gap as \emph{conditional bias}. Defining the per-event bias $\mathbf{b}(\mathbf{x}, t; e) \triangleq \mathbf{s}^\star(\mathbf{x}, t \mid \mathbf{h}_{e-1}, \mathbf{c}_e) - \bar{\mathbf{s}}^\star(\mathbf{x}, t \mid \mathbf{c}_e)$, the resulting gradient decomposes as
\begin{equation}
    \nabla_\theta \mathcal{L}^{\mathrm{biased}}_{\mathrm{DMD}}
    =
    -\mathbb{E}\!\left[
    \big(\bar{\mathbf{s}}^\star - \mathbf{s}_{\mathrm{fake}}\big)
    \frac{\partial G_\theta}{\partial \theta}
    \right]
    -
    \textcolor{blue}{
    \underbrace{
    \mathbb{E}\!\left[
    \mathbf{b}(\mathbf{x}, t; e)
    \frac{\partial G_\theta}{\partial \theta}
    \right]
    }_{\text{conditional bias}}
    },
\end{equation}
where the first term is the nominal DMD update against the marginalized teacher and the second is the spurious direction induced by the history--condition mismatch.

\noindent\textbf{Motivation: trust regions for unreliable teachers.} Temporal incoherence is therefore not a capacity or optimization issue, but a structural mismatch in supervision: teacher guidance that is not reliably trajectory-conditioned cannot be trusted uniformly across events. A very direct solution is therefore to accept a teacher update only when it is consistent with the generator's established trajectory. We draw inspiration from TRPO~\cite{schulman2015trust}, which constrains policy updates to a region where the underlying advantage estimate remains trustworthy. This trust-region principle motivates the reliability-aware design of \framework, detailed in Sec.~\ref{sec:method}.

%% file: sec/4_method.tex
\section{Method}
\label{sec:method}

\noindent\textbf{From principle to design.} The trust-region principle from Sec.~\ref{sec:motivation} prescribes \emph{when} to trust the teacher, but not \emph{what} to do when the teacher cannot be trusted. We resolve this by exploiting an asymmetry between the two scores in DMD: $\mathbf{s}_{\mathrm{real}}$ is \emph{frozen} and, once biased, persistently pulls $G_\theta$ off-trajectory; $\mathbf{s}_{\mathrm{fake}}$ is \emph{dynamically updated} on samples from $G_\theta$~\cite{yin2024improved} and therefore accumulates momentum reflecting the actual generation history. This suggests a two-part design: \emph{adaptively modulate} the real score's influence by its reliability, and \emph{fall back} on the fake trajectory's own momentum which is bias-free by construction when reliability is low. We thus formulate the \framework objective as
\begin{equation}
    \mathcal{L} = w_k\,\mathcal{L}_{\mathrm{DMD}}
               + (1 - w_k)\,\mathcal{L}_{\mathrm{cont}},
    \label{eq:final}
\end{equation}

where $w_k \in (0,1)$ is an online reliability weight realizing the trust-region radius and $\mathcal{L}_{\mathrm{cont}}$ is a bias-immune continuity loss grounded in the fake trajectory's momentum. Under this formulation, two questions remain: \emph{(i)} how to define $\mathcal{L}_{\mathrm{cont}}$ to faithfully capture this momentum, and \emph{(ii)} how to estimate $w_k$ online to detect when the real score is biased. We address each in turn.

\noindent\textbf{Observation 1.} \textit{The fake trajectory carries a history-grounded, bias-free signal that the real score lacks.} Following the chunk-wise autoregressive paradigm of LongLive~\cite{yangLongLiveRealtimeInteractive2025}, the generator produces fake latent chunks
\begin{equation}
    \mathbf{x}^{(k)}_{\mathrm{fake}} \sim G_\theta\!\left(C_k,\;
    \mathbf{c}_{e(k)}\right),
\end{equation}
where $C_k$ is the accumulated KV cache encoding the trajectory history $\mathbf{h}_{e(k)-1}$ and $e(k)$ is the event index at chunk $k$. Because $C_k$ is built entirely from the generator's own past outputs, the chunk-to-chunk evolution of $\mathbf{x}^{(k)}_{\mathrm{fake}}$ encodes where generation has been independently of what the frozen real score advocates. This is precisely the fall-back signal anticipated in Eq.~\eqref{eq:final}: bias-free by construction, and informative exactly when $\mathbf{s}_{\mathrm{real}}$ is not.

\noindent\textbf{Design 1: Monotonic improvement via continuity loss on the fake trajectory.} To turn this momentum into a usable supervision signal, we summarize each fake chunk by a semantic descriptor via a frozen DINO~\cite{oquab2023dinov2,simeoni2025dinov3} feature extractor $\Phi$ and penalize abrupt deviations between consecutive descriptors:
\begin{equation}
    \mathcal{L}_{\mathrm{cont}} =
    \left\|\mathbf{f}^{\mathrm{fake}}_k -
    \mathbf{f}^{\mathrm{fake}}_{k-1}\right\|_2^2,
    \quad
    \mathbf{f}^{\mathrm{fake}}_k =
    \Phi\!\left(\mathbf{x}^{(k)}_{\mathrm{fake}}\right).
    \label{eq:cont}
\end{equation}
$\mathcal{L}_{\mathrm{cont}}$ involves neither $\mathbf{s}_{\mathrm{real}}$ nor $\hat{\mathbf{x}}^{(k)}_{\mathrm{real}}$, and is therefore immune to $\mathbf{b}(\mathbf{x}, t; e)$ by construction. Rather than prescribing where the fake trajectory should go, it preserves the trajectory's own momentum by penalizing deviations from its prior evolution, realizing the ``fall-back'' branch of our objective.

\input{tables/algorithm}
\noindent\textbf{Observation 2.} \textit{Real score reliability is observable: bias surfaces as abrupt semantic jumps in the teacher's chunk-to-chunk evolution.} Empirically, conditional bias does not manifest as uniform noise but as sharp discontinuities: the teacher's denoised estimate $\hat{\mathbf{x}}^{(k)}_{\mathrm{real}}$ shifts smoothly across most chunks but jumps suddenly when the new condition $\mathbf{c}_e$ overrides the trajectory history. Between such jumps, the teacher's evolution remains consistent with the student's history-grounded fake trajectory. At a jump, the two diverge sharply. Real score reliability is therefore not a hidden quantity to be inferred from training dynamics, but a \emph{locally observable} property of the teacher's chunk-level evolution, reducing online estimation of $w_k$ to detecting when this evolution becomes non-smooth.

\noindent\textbf{Design 2: Adaptive trust region from latent-delta discrepancy.} Observation 2 turns $w_k$ from an abstract reliability weight into a concrete quantity: a measure of how closely the teacher's chunk-to-chunk evolution tracks the student's. We instantiate it in two steps (i) measure this evolution discrepancy in a semantic feature space, and (ii) translate it into the trust-region radius of Eq.~\eqref{eq:final}.

\textit{Smoothness measurement.} At each chunk $k$, we encode both $\mathbf{x}^{(k)}_{\mathrm{fake}}$ and $\hat{\mathbf{x}}^{(k)}_{\mathrm{real}}$ with the DINO~\cite{oquab2023dinov2,simeoni2025dinov3} feature extractor $\Phi$ and take consecutive differences across events:
\begin{align}
    \boldsymbol{\delta}^{\mathrm{fake}}_k &=
        \Phi\!\left(\hat{\mathbf{x}}^{(k)}_{\mathrm{fake}}\right) -
        \Phi\!\left(\hat{\mathbf{x}}^{(k-1)}_{\mathrm{fake}}\right), \\
    \boldsymbol{\delta}^{\mathrm{real}}_k &=
        \Phi\!\left(\hat{\mathbf{x}}^{(k)}_{\mathrm{real}}\right) -
        \Phi\!\left(\hat{\mathbf{x}}^{(k-1)}_{\mathrm{real}}\right).
    \label{eq:deltas}
\end{align}
which serve as the empirical counterparts of $\Delta\mathbf{s}^\star_e$ and $\Delta\mathbf{s}_{\mathrm{real}}^e$ from Sec.~\ref{sec:motivation}. The DINO feature space exposes the abrupt jumps identified in Observation 2 while remaining insensitive to pixel-level noise.

\textit{Trust-region gating.} We let the discrepancy between the two deltas define the radius of a trust region in latent trajectory space, within which the DMD gradient is permitted to steer the generator:
\begin{equation}
    \rho_k = \left\|\boldsymbol{\delta}^{\mathrm{real}}_k -
    \boldsymbol{\delta}^{\mathrm{fake}}_k\right\|_2,
    \qquad
    w_k = \sigma\!\left(-(\rho_k - \mu)\cdot s\right),
    \label{eq:trust}
\end{equation}


where $\mu$ and $s$ control the detection threshold and the sharpness of the transition. When $\rho_k$ is small, the teacher evolves in step with the student's history, the trust region is wide ($w_k \to 1$), and the DMD update proceeds in full. When $\rho_k$ spikes, conditional bias is active, the trust region collapses ($w_k \to 0$), and supervision falls back on $\mathcal{L}_{\mathrm{cont}}$. Unlike classical trust regions defined in parameter space, ours lives directly in the observable latent trajectory space: it requires no second-order gradient information and is computable online from a single feature-space distance, making it a lightweight realization of the ``adaptive modulation'' branch of our objective.

\noindent\textbf{\framework.} Substituting Designs 1 and 2 into Eq.~\eqref{eq:final} gives the complete \framework objective, as in Alg.~\ref{alg:deltaforcing}. The two branches, \emph{(i)} adaptive modulation of the real score and \emph{(ii)} fall-back on the fake trajectory's momentum, are now realized by $w_k$ and $\mathcal{L}_{\mathrm{cont}}$, respectively, and coordinate through a single online quantity $\rho_k$. Within a stable event, $\rho_k$ is small, $w_k \to 1$, and training reduces to standard DMD. When the frozen real score shifts abruptly, the fake trajectory continues its prior momentum, $\rho_k$ spikes, $w_k \to 0$, and $\mathcal{L}_{\mathrm{cont}}$ takes over to preserve that momentum. As the new condition stabilizes, real and fake deltas realign, $\rho_k$ subsides, and the DMD term gradually reasserts itself.

%% file: tables/algorithm.tex
\begin{wrapfigure}{r}{0.5\linewidth}
\begin{minipage}{\linewidth}
\begin{algorithm}[H]
\caption{\framework}
\scriptsize
\label{alg:deltaforcing}
\begin{algorithmic}[1]
\Require Causal video generator $G_\theta$, Prompt set $\mathcal{P}$
\Require Video length $l_{\text{video}}$, Chunk length $l_{\text{chunk}}$
\Require $\Phi$, $\mu$, $s$
\While{not converged}
    \State Initialize KV cache $C \gets []$
    \State Initialize current video length $l \gets 0$
    \State Sample $(p,\ p_{\text{next}})\sim\mathcal{P}$
    \State Sample switch index $\tau$
    \State $\tau \gets \tau \cdot l_{\text{chunk}}$

    \If{$l \ge l_{\text{video}}$}
      \State $C \gets []$;\quad $l \gets 0$
      \State Resample $(p,\ p_{\text{next}})$ and $\tau$
    \EndIf

    \State $p_{\text{active}} \gets
    \begin{cases}
        p, & l < \tau \\
        p_{\text{next}}, & \text{otherwise}
    \end{cases}$

    \If{$l = \tau$}
      \State $C \gets \texttt{recache}(G_\theta,\ \mathbf{v},\ C,\
      p_{\text{active}})$
    \EndIf
    \State $\mathbf{x}^{(k)} \gets \texttt{generate\_next\_chunk}(G_\theta,\ C,\ p_{\text{active}})$
    \State \gitrem{$\mathcal{L} \gets \texttt{DMD\_Loss}(G_\theta,\
        \mathbf{x}^{(k)},\ p_{\text{active}})$}
    \State \gitadd{$(\hat{\mathbf{x}}^{(k)}_{\mathrm{real}},\
        \mathcal{L}_{\mathrm{DMD}}) \gets \texttt{DMD\_Loss}(G_\theta,\
        \mathbf{x}^{(k)},\ p_{\text{active}})$}
    
    \State \gitadd{$\mathbf{f}^{\mathrm{fake}}_k \gets
        \Phi\!\left(\mathbf{x}^{(k)}_{\mathrm{fake}}\right)$}
    \State \gitadd{$\mathbf{f}^{\mathrm{real}}_k \gets
        \Phi\!\left(\hat{\mathbf{x}}^{(k)}_{\mathrm{real}}\right)$}

    \State \gitadd{$\boldsymbol{\delta}^{\mathrm{fake}}_k \gets
        \mathbf{f}^{\mathrm{fake}}_k - \mathbf{f}^{\mathrm{fake}}_{k-1}$}
    \State \gitadd{$\boldsymbol{\delta}^{\mathrm{real}}_k \gets
        \mathbf{f}^{\mathrm{real}}_k - \mathbf{f}^{\mathrm{real}}_{k-1}$}

    \State \gitadd{$\rho_k \gets
        \|\boldsymbol{\delta}^{\mathrm{real}}_k -
        \boldsymbol{\delta}^{\mathrm{fake}}_k\|_2$}
    \State \gitadd{$w_k \gets
        \sigma\!\left(-(\rho_k - \mu)\cdot s\right)$}

    \State \gitadd{$\mathcal{L}_{\mathrm{cont}} \gets
        \|\mathbf{f}^{\mathrm{fake}}_k -
        \mathbf{f}^{\mathrm{fake}}_{k-1}\|_2^2$}
    \State \gitadd{$\mathcal{L} \gets
        w_k\,\mathcal{L}_{\mathrm{DMD}} +
        (1-w_k)\,\mathcal{L}_{\mathrm{cont}}$}

    \State $\mathcal{L}.\texttt{backward}()$
    \State update generator parameter $\theta$
    \State $l \gets l + l_{\text{chunk}}$
\EndWhile
\end{algorithmic}
\end{algorithm}
\end{minipage}
\vspace{-2em}
\end{wrapfigure}

%% file: sec/5_experiments.tex
\section{Experiments}
\label{sec:exp}

\subsection{Implementation Details}

\noindent\textbf{Implementation.}
We build \framework on top of LongLive~\cite{yangLongLiveRealtimeInteractive2025}. The generator is WAN-2.1-1.3B-T2V~\cite{wan2025wan} and the teacher is WAN-2.1-14B-T2V~\cite{wan2025wan}, which natively produces 5-second clips at 16 FPS and $832\times480$ resolution. Training proceeds in two stages as LongLive~\cite{yangLongLiveRealtimeInteractive2025}. In Stage 1, we replace the original Self Forcing~\cite{huangSelfForcingBridging2025} initialization with Causal Forcing~\cite{zhuCausalForcingAutoregressive2026} for the context window and frame sink, and train for 700 steps with learning rates of $2\times10^{-6}$ for the generator and $4\times10^{-7}$ for $\mathtt{S_{fake}}$. In Stage 2, we apply \framework together with \textit{Streaming Long Tuning} for an additional 3{,}000 steps, with learning rates of $1\times10^{-5}$ and $2\times10^{-6}$, respectively. All of our experiments are conducted on Nvidia H100.

\noindent\textbf{Baselines.}
We compare \framework against two families of methods at a comparable scale: open-source causal video generators, including SkyReels-V2~\cite{chen2504skyreels} and MAGI-1~\cite{teng2025magi}, and distilled causal video generators including LongLive~\cite{yangLongLiveRealtimeInteractive2025}, MemFlow~\cite{jiMemFlowFlowingAdaptive2025}, and Reward Forcing~\cite{luRewardForcingEfficient2025}. For a fair comparison, all distilled causal baselines are initialized following the Causal Forcing~\cite{zhuCausalForcingAutoregressive2026}, and Reward Forcing is further trained under \textit{Streaming Long Tuning} to enhance interactive generation capability.

\noindent\textbf{Evaluation.}
We evaluate \framework along four complementary axes. \ding{202} We adopt VBench~\cite{huang2023vbench,zheng2025vbench2,huang2025vbench++} to assess overall performance. Specifically, we adopt six dimensions: Subject Consistency, Background Consistency, Aesthetic Quality, Motion Smoothness, Dynamic Degree, and Imaging Quality. \ding{203} We measure instruction-following ability using Long-CLIP~\cite{zhang2024longclip}. \ding{204} We further evaluate human-preference alignment with VideoAlign~\cite{liu2025improving}, which scores visual quality (VQ), motion quality (MQ), and text alignment (TA). \ding{205} Finally, we conduct a user study to obtain direct human preference evaluation. All experiments are conducted on the MemFlow benchmark~\cite{jiMemFlowFlowingAdaptive2025}, which contains 100 sequences, each consisting of six sequential 10-second events and forming a 60-second video.

\subsection{Quantitative Results}

\noindent\textbf{\ding{202} Results on VBench~\cite{huang2023vbench}.}

\input{tables/vbench}
As shown in Table~\ref{tab:vbench}, \framework attains the best subject and background consistency within the distilled-causal group while matching competing methods on motion smoothness, aesthetic quality, and imaging quality. The large gains on transition-sensitive consistency without trading off perceptual or motion quality indicates that the improvements come from more stable cross-event trajectory evolution under repeated condition updates rather than from a shift along any single perceptual axis. Methods optimized for instantaneous responsiveness remain highly reactive but accumulate transition mismatch over long horizons. Our trust-region supervision directly targets this trade-off by suppressing trajectory-incompatible teacher updates across events.

\noindent\textbf{\ding{203} \framework maintains instruction-following capability.}
Since commonly adopted prompt-enhancement strategies~\cite{wan2025wan,wu2025hunyuanvideo} typically produce prompts that are more detailed and substantially longer than standard CLIP-style captions~\cite{clip}, we measure prompt--video alignment for each interaction chunk using Long-CLIP~\cite{zhang2024longclip}. This metric directly evaluates whether the model can follow newly arriving prompts as the conditioning signal evolves over time. As shown in Table~\ref{tab:alignment}, \framework remains consistently top-ranked or near top-ranked from early to late windows, indicating that transition-aware supervision preserves responsiveness while improving long-horizon stability.

\input{tables/alignment}

\noindent\textbf{\ding{204} \framework achieves stronger human preference.}
We further evaluate \framework using VideoAlign~\cite{liu2025improving}, a reward model trained from human feedback, to better capture perceptual quality and human preference in interactive generation. To make the evaluation boundary-sensitive, we concatenate each generated interaction chunk with the final 1-second segment of the preceding interaction and score the resulting stitched clip. This protocol explicitly emphasizes event continuity, where conditional bias is most likely to induce drift. As shown in Table~\ref{tab:alignment}, \framework achieves the best overall VideoAlign score, with the highest motion quality and text alignment, while remaining competitive in visual quality.

\input{tables/user_study}
\noindent\textbf{\ding{205} User Study.}
We conduct a side-by-side user study with 20 participants to assess quality in multi-event streaming generation. In each trial, participants view four videos generated under the same sequence of event-driven prompts and rank them along three criteria: aesthetic quality, dynamic quality, and multi-event naturalness. Because this ranking is cognitively demanding, each participant completes only three trials, randomly sampled from the full set. Video order within each trial is randomized. All methods are evaluated on identical prompt schedules, enabling direct comparison of both responsiveness to condition changes and temporal consistency across event boundaries. The user-study interface is shown in Appendix~\ref{apx:userstudy}. As summarized in Table~\ref{tab:user_study}, \framework attains the best (lowest) average rank across all three criteria, indicating a clear human preference. The margin is largest on dynamic quality and multi-event naturalness, consistent with our method's design goal of preserving coherent motion and producing smoother transitions under evolving conditions.

\begin{figure}[t]
    \centering
    \includegraphics[width=0.99\linewidth]{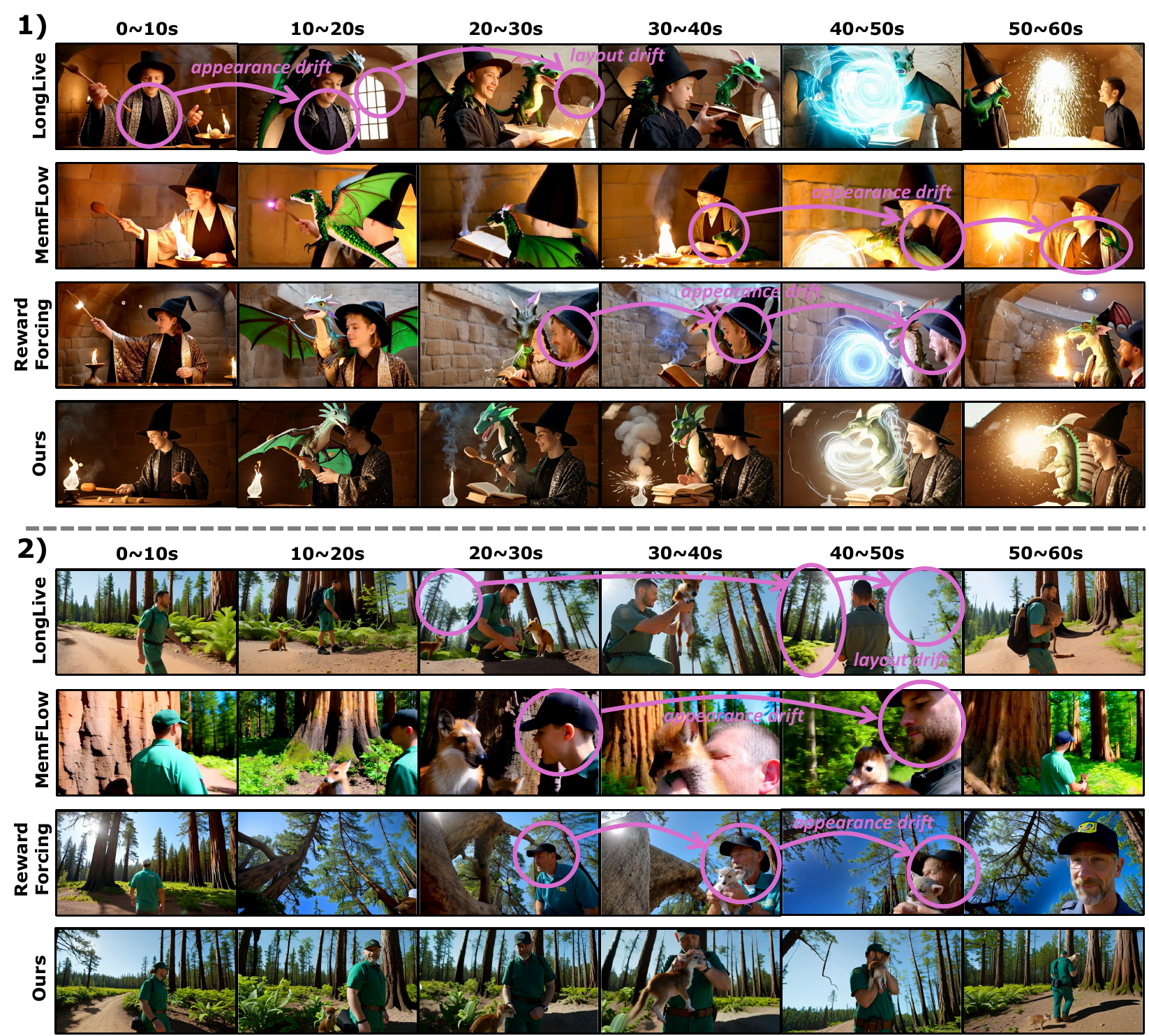}
    \caption{Qualitative results. Each 10s segment corresponds to one event and the full event prompts are listed in Appendix~\ref{apx:exp-details}.}
    \label{fig:visualization}
    \vspace{-1em}
\end{figure}

\subsection{Qualitative Results}

Figure~\ref{fig:visualization} presents qualitative comparisons for multi-event generation. The regions highlighted by purple circles reveal the failure patterns in existing methods: the generated content drifts across events, manifesting as changes in object identity, scene layout, and overall structure. This clearly indicates that the trajectory is not smoothly evolved, but instead repeatedly redirected toward different modes after each event. Although MemFlow~\cite{jiMemFlowFlowingAdaptive2025} incorporates memory mechanisms that can partially recover from drift (\eg, in Fig.~\ref{fig:visualization}, Example 1, the appearance returns closer to the initial state from 40--50s to 50--60s), it still fails to maintain consistent trajectories due to biased supervision. This suggests that memory alone is insufficient when the teacher guidance itself is not history-aware. In contrast, \framework, despite being built upon a memoryless generator following LongLive~\cite{yangLongLiveRealtimeInteractive2025}, maintains consistent object identity, structure, and motion across events. The generated trajectory evolves smoothly without abrupt deviations, demonstrating that our method effectively suppresses biased teacher guidance and preserves trajectory-consistent dynamics. We further provide latent-trajectory visualizations in Appendix~\ref{apx:vis-latent-traj} to empirically validate \framework's effectiveness.

%% file: tables/vbench.tex
\begin{wraptable}{r}{0.63\textwidth}
  \scriptsize
  \setlength{\tabcolsep}{2pt} 
  \vspace{-4em}
  \caption{
    Comparisons with related representative open-source autoregressive video generation models of similar parameter sizes on VBench~\cite{huang2023vbench}.
  }
  \label{tab:vbench}
  \centering
  \renewcommand{\arraystretch}{1.3}
  \resizebox{\linewidth}{!}{%
  \begin{tabular}{>{\raggedright\arraybackslash}p{0.24\linewidth}c|cccccc}
      \toprule
      \multirow{3}{*}{Model} & \multirow{3}{*}{\#Params} &
      \multicolumn{6}{c}{Evaluation Scores $\uparrow$}  \\
    \cmidrule(lr){3-8}
      && \scriptsize Subject       & \scriptsize Background    & \scriptsize Motion    & \scriptsize Aesthetic & \scriptsize Imaging & \scriptsize Dynamic \\
      && \scriptsize Consistency   & \scriptsize Consistency   & \scriptsize Smoothness& \scriptsize Quality   & \scriptsize Quality & \scriptsize Degree \\
    \midrule
    \rowcolor[HTML]{EEE5F4}
    \multicolumn{8}{l}{\textit{Diffusion Forcing Causal}}\\
    SkyReels-V2~\citep{chen2504skyreels} & 1.3B & 91.64 & 92.88 & 98.46 & 56.97 & 64.33 & 88.37 \\
    MAGI-1~\citep{teng2025magi}      & 4.5B & 92.63 & 91.20 & 98.43 & 59.18 & 65.12 & 82.04 \\
    \midrule
    \rowcolor[HTML]{EEE5F4}
    \multicolumn{8}{l}{\textit{Distilled Causal}}\\
    LongLive~\cite{yangLongLiveRealtimeInteractive2025} & 1.3B & 94.97 & 93.37 & 98.37 & 58.95 & 72.59 & 90.93 \\
    MemFlow~\cite{jiMemFlowFlowingAdaptive2025} & 1.3B & 93.19 & 92.03 & 97.17 & 55.91 & 69.57 & 91.71 \\
    Reward Forcing~\citep{luRewardForcingEfficient2025}       & 1.3B & \underline{95.55} & \underline{93.40} & \underline{98.51} & 56.32 & \textbf{74.99} & \textbf{93.16} \\
    \rowcolor[rgb]{0.9,0.95,1}
    Delta Forcing (ours) & 1.3B & \textbf{96.60} & \textbf{94.63} & \textbf{98.78} & \textbf{58.98} & \underline{72.72} & \underline{92.18} \\
    \bottomrule
  \end{tabular}
  }
\end{wraptable}

%% file: tables/alignment.tex
\begin{table}[t]
  \footnotesize
  \setlength{\tabcolsep}{4pt} 
  \vspace{-2em}
  \caption{
    Comparisons with Long-CLIP Score~\cite{clip} and VideoAlign~\cite{liu2025improving} metrics across different video generation baselines.
  }
  \label{tab:alignment}
  \centering
  \resizebox{\textwidth}{!}{ 
  \renewcommand{\arraystretch}{1.5}
  \begin{tabular}{l|ccccccc|cccc}
    \toprule
    \multirow{2}{*}{Model} & \multicolumn{7}{c|}{Long-CLIP Score $\uparrow$} & \multicolumn{4}{c}{VideoAlign $\uparrow$} \\
    \cmidrule(lr){2-8} \cmidrule(lr){9-12}
    & 0-10s & 10-20s & 20-30s & 30-40s & 40-50s & 50-60s & \textbf{Avg.} & VQ & MQ & TA & \textbf{Total} \\
    \midrule
    \rowcolor[HTML]{EEE5F4}
    \multicolumn{12}{l}{\textit{Diffusion Forcing Causal}}\\
    SkyReels-V2~\citep{chen2504skyreels} & 27.08 & 25.29 & 24.42 & 24.25 & 23.88 & 23.79 & 24.79 & 2.24 & 3.45 & -0.20 & 5.49 \\
    MAGI-1~\citep{teng2025magi}      & 27.15 & 25.25 & 23.73 & 23.67 & 23.40 & 23.61 & 24.47 & 1.95 & 1.53 & -0.83 & 2.66 \\
    \midrule
    \rowcolor[HTML]{EEE5F4}
    \multicolumn{12}{l}{\textit{Distilled Causal}}\\
    LongLive~\cite{yangLongLiveRealtimeInteractive2025}       & \underline{27.69} & 26.39 & \underline{25.68} & \underline{25.53} & \textbf{25.35} & \textbf{25.66} & \underline{26.05} & 2.67 & 3.96 & 0.50 & 7.13 \\
    MemFlow~\cite{jiMemFlowFlowingAdaptive2025}          & 27.27 & \underline{26.47} & 25.44 & 25.36 & \underline{25.27} & 25.13 & 25.82 & \textbf{2.83} & 4.10 & \underline{0.54} & 7.46 \\
    Reward Forcing~\citep{luRewardForcingEfficient2025}      & 27.58 & 25.04 & 24.35 & 23.95 & 23.80 & 23.75 & 24.75 & 2.72 & \underline{4.18} & 0.04 & 6.94 \\
    \rowcolor[rgb]{0.9,0.95,1} 
    \textbf{Delta Forcing (Ours)} & \textbf{28.13} & \textbf{26.48} & \textbf{25.70} & \textbf{25.58} & 25.25 & \underline{25.26} & \textbf{26.07} & \underline{2.75} & \textbf{4.18} & \textbf{0.61} & \textbf{7.55} \\
    \bottomrule
  \end{tabular}
  }
\end{table}

%% file: tables/user_study.tex
\begin{wraptable}{r}{0.55\textwidth}
    \footnotesize
    \setlength{\tabcolsep}{1.5pt}
    \vspace{-1em}
    \caption{
      \textbf{User Study.} Evaluation results on human preference via average rank.
    }
    \label{tab:user_study}
    \centering
    \renewcommand{\arraystretch}{1.3}
    \resizebox{\linewidth}{!}{%
    \begin{tabular}{l|cccc}
      \toprule
      \textbf{Model} & \textbf{Aesthetic} $\downarrow$ & \textbf{Dynamic} $\downarrow$ & \textbf{Naturalness}$^{*}$ $\downarrow$ & \textbf{Average} $\downarrow$ \\
      \midrule
      LongLive       & 2.83 & 2.39 & 2.56 & 2.59 \\
      MemFlow        & 2.61 & 3.17 & 2.94 & 2.91 \\
      Reward Forcing & 2.39 & 2.50 & 2.72 & 2.54 \\
      \rowcolor[rgb]{0.9,0.95,1}
      \textbf{Delta Forcing (Ours)} & \textbf{2.17} & \textbf{1.94} & \textbf{1.78} & \textbf{1.96} \\
      \bottomrule
    \end{tabular}
    }
    \vspace{0.2em}
    \flushleft
    \scriptsize{$^{*}$Naturalness is evaluated from a multi-event perspective, measuring the smoothness and coherence of transitions across events.}
    \vspace{-1em}
\end{wraptable}

%% file: sec/6_discussion.tex
\section{Ablation Studies}
\label{sec:ablation}

\phantom{x}
\begin{wrapfigure}{r}{0.5\textwidth}
  \vspace{-6em}
  \begin{center}
    \includegraphics[width=\linewidth]{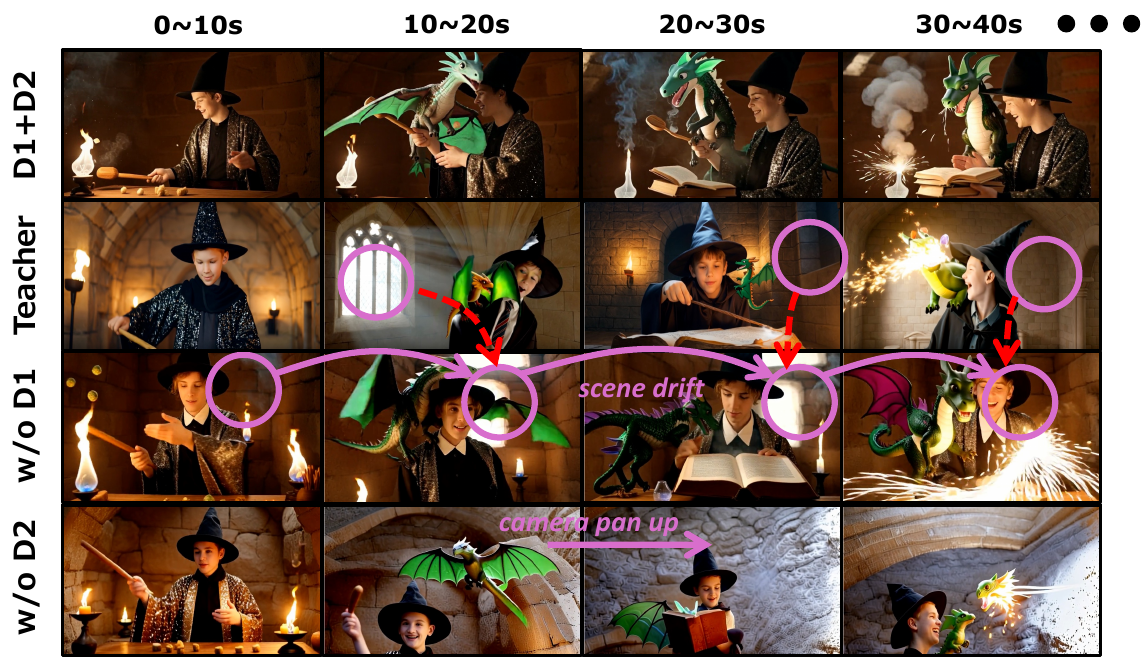}
  \end{center}
  \caption{Ablation study.}
  \label{fig:ablation}
  \vspace{-1em}
\end{wrapfigure}
\noindent\textbf{Without continual loss (Design 1).}
In this experiment, we remove the monotonic continual loss, \ie, $(1-w_k)\mathcal{L}_{\mathrm{cont}}$, and only retain the adaptive trust-region regulation on the DMD objective. As shown in Fig.~\ref{fig:ablation}(row 2), the object appearance and local motion remain relatively consistent, suggesting that the trust-region mechanism can still suppress part of the unreliable teacher guidance. However, the global scene layout gradually drifts during generation. This indicates that, without the continual loss, the model lacks an explicit constraint to anchor its current prediction to the accumulated trajectory, and thus can still be biased by condition-aligned but trajectory-agnostic teacher supervision.

\noindent\textbf{Without adaptive trust regions (Design 2).}
We then remove the adaptive trust-region weight $w_k$ from the original DMD loss, so that teacher supervision is no longer selectively suppressed according to its reliability. As shown in Fig.~\ref{fig:ablation}(row 3), the camera consistently pans upward, revealing a typical mode-seeking failure in AR-DiT distillation. We hypothesize that this degeneration arises because biased DMD gradients are applied without normalized reliability control, allowing certain motion directions to be excessively amplified and eventually dominate the generation trajectory. Similar failures are also observed in Reward Forcing~\cite{luRewardForcingEfficient2025}, where dynamics-oriented reward scaling can over-strengthen motion-related gradients and lead to persistent camera panning in multi-event settings. This suggests that the original DMD objective should not be uniformly encouraged under evolving conditions. Instead, it must be regulated by normalized adaptive weights to prevent unreliable teacher guidance, consistent with recent findings~\cite{jiang2025dmdr,bai2026amd}.

%% file: sec/7_conclusion.tex
\section{Conclusion}
\label{sec:conclusion}

We identify \emph{conditional bias} as a structural failure mode in distillation-based streaming video generation: the teacher is condition-aware but trajectory-agnostic, so distribution-matching supervision repeatedly steers the student toward condition-consistent but trajectory-inconsistent modes, accumulating drift across event boundaries. We propose \framework, a reliability-aware framework inspired by trust-region policy optimization that compares each teacher-induced transition with the generator's history-grounded evolution, reinforcing supervision within a trajectory-consistent trust region and suppressing guidance that deviates from it. Extensive experiments show that \framework substantially improves temporal stability while preserving event reactivity in long-horizon real-time interactive multi-event video generation.

%% file: sec/appendix.tex
\section{Motivation Study via Latent Trajectory Visualization}
\label{apx:motivation}

To supplement our motivation analysis, we provide a latent-space diagnostic that reveals how existing interactive streaming video generation methods behave under evolving event conditions. Concretely, we extract the denoised latent feature of every generated frame prior to VAE decoding and visualize the temporal trajectory of these latent states in a two-dimensional PCA space. This analysis offers direct evidence of whether the generation process evolves smoothly across prompt switches or instead exhibits drift, collapse, or mode-seeking behavior.

\subsection{What Should a Good Trajectory Look Like?}
\label{sec:desirable}

Before presenting the visualization protocol and results, we first establish the criteria against which trajectories should be evaluated. From the perspective of interactive multi-event video generation, a well-behaved latent trajectory should satisfy two complementary properties:

\begin{itemize}
    \item \textbf{Within-interaction aggregation.} Frames belonging to the same prompt interval should form a compact local cluster, reflecting stable identity, scene layout, and motion style. Such aggregation indicates that the model maintains a coherent visual state while generating temporally continuous frames under a fixed condition.
    
    \item \textbf{Smooth cross-interaction displacement.} When the prompt changes, the trajectory should move from the previous local region toward a new one in a gradual and structured manner. The displacement should be sufficient to reflect the semantic shift, yet not so abrupt as to break temporal continuity.
\end{itemize}

In other words, a desirable trajectory should resemble a sequence of compact local groups connected by smooth bridges. This structure reflects the central trade-off of our problem: within-interaction stability and between-interaction responsiveness must coexist. Excessively small displacement implies insufficient adaptation, whereas abrupt jumps indicate loss of temporal consistency. With these criteria in mind, we now describe the visualization protocol and examine whether existing methods satisfy them.

\subsection{Visualization Implementations}
\label{sec:protocol}

\paragraph{Latent extraction and projection.}
For each model, we generate a video under the same multi-event interaction setting in which the prompt changes at fixed intervals. Suppose the video contains $T$ frames divided into $K$ interaction segments, with each segment corresponding to one sub-prompt. Let $\mathbf{z}_t \in \mathbb{R}^d$ denote the denoised latent representation of frame $t$ before VAE decoding. We collect the full latent sequence:
\begin{equation}
\mathcal{Z} = \{\mathbf{z}_1, \mathbf{z}_2, \dots, \mathbf{z}_T\}.
\end{equation}

We fit PCA on $\mathcal{Z}$ and project each frame-level latent into a two-dimensional space:
\begin{equation}
\tilde{\mathbf{z}}_t = \mathbf{W}^\top (\mathbf{z}_t - \boldsymbol{\mu}), \qquad \tilde{\mathbf{z}}_t \in \mathbb{R}^2,
\end{equation}
where $\boldsymbol{\mu}$ is the empirical mean of the sequence and $\mathbf{W} \in \mathbb{R}^{d \times 2}$ contains the top two principal directions. The projected points $\{\tilde{\mathbf{z}}_t\}_{t=1}^T$ are connected in temporal order to form a trajectory, with different colors denoting different interaction segments. We additionally annotate the start and end of selected segments to facilitate reading of transition patterns. To ensure fair comparison, all methods share the same prompt schedule, switching timestamps, and latent extraction stage.

\paragraph{Choice of projection method.}
We adopt PCA rather than t-SNE~\cite{tsne} or UMAP~\cite{2018arXivUMAP} because our objective is not cluster discovery but \emph{temporal trajectory analysis}. This distinction has three concrete implications. \textit{First}, PCA preserves the \emph{global geometry} of the latent space. As a linear projection along the directions of maximal variance, it maintains relative distances and directional relationships. t-SNE and UMAP prioritize local neighborhood fidelity and can substantially distort global arrangement, which is acceptable for cluster visualization but problematic when assessing whether a sequence follows a coherent trajectory. \textit{Second}, PCA faithfully reflects \emph{transition characteristics}. Because it preserves large-scale directional variation, the projected trajectory allows us to inspect transition magnitude, smoothness, and cross-event displacement without the artificial stretching or compression that nonlinear methods may introduce. A smooth trajectory that appears fragmented, or an abrupt jump that appears mild, would directly undermine our diagnostic purpose. \textit{Third}, PCA is \emph{deterministic and reproducible}. t-SNE and UMAP are sensitive to perplexity, neighborhood size, initialization, and random seed. Since our conclusions depend on comparing trajectory patterns across methods under identical conditions, we require a projection whose behavior is stable across runs.

\begin{figure*}[t]
    \centering
    \includegraphics[width=\textwidth]{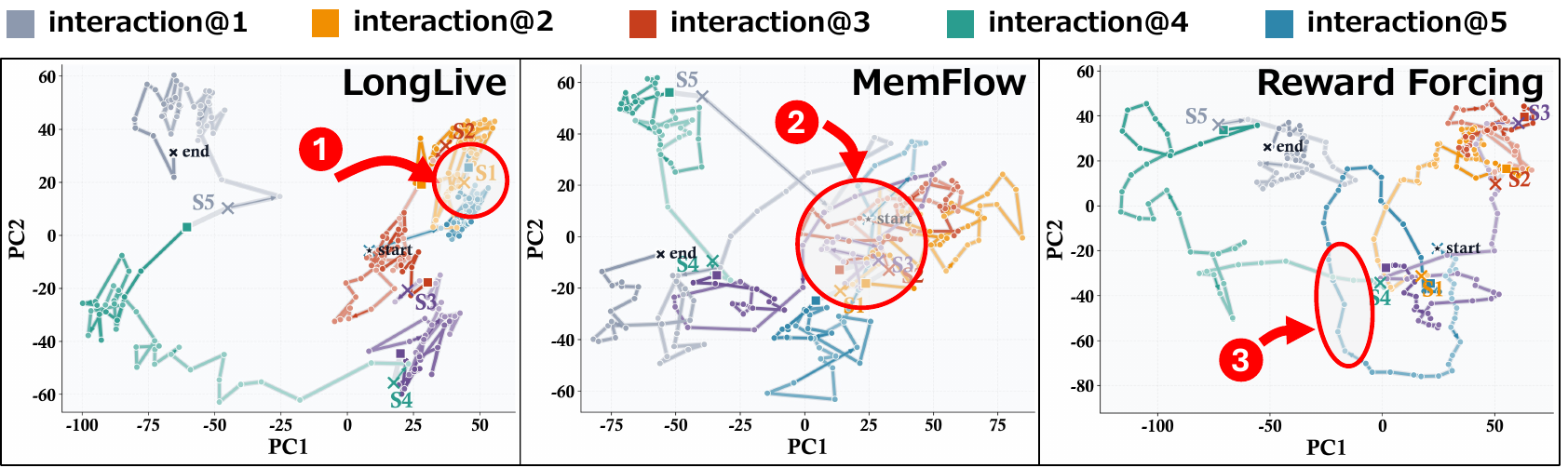}
    \caption{\textbf{Latent trajectory visualization via PCA under multi-event prompt switching.}
    We project frame-wise denoised latent features (before VAE decoding) into a two-dimensional PCA space and connect them in temporal order. Different colors denote different interaction segments. Left exhibits short and narrow transitions across prompt switches, indicating insufficient semantic displacement despite changed conditions. Middle: shows irregular and scattered patterns with no clear temporal structure, indicating severe inter-event drift. Right evolves with persistent motion regardless of interaction boundaries, suggesting mode-seeking dynamics in which within-event and cross-event transitions become indistinguishable. Comparing against the desirable properties defined in Section~\ref{sec:desirable}, none of these methods achieves both within-interaction aggregation and smooth cross-interaction displacement.}
    \label{fig:latent_pca}
\end{figure*}

\subsection{Empirical Observations}
\label{sec:observations}

Figure~\ref{fig:latent_pca} presents the PCA trajectories of three representative methods under consecutive prompt switching. We evaluate each against the desirable properties established in Section~\ref{sec:desirable}.

\textbf{\textcolor{red}{\ding{182}} Under-reactive transitions.}
The trajectory exhibits short, concentrated transitions across the first several prompt segments. Despite clear shifts in the semantic condition, the latent path displaces only modestly, remaining anchored to preceding states rather than moving toward the new event condition. With respect to our criteria, this method partially satisfies within-interaction aggregation but fails on cross-interaction displacement: the trajectory appears temporally smooth, yet the displacement is too weak to reflect the actual semantic change. This pattern indicates an under-reactive transition mode in which continuity is preserved at the cost of insufficient adaptation.

\textbf{\textcolor{red}{\ding{183}} Unstructured drift.}
The trajectory exhibits a markedly different failure. Rather than forming a structured progression, latent states scatter across the projected space without directional regularity. Consecutive prompt transitions correspond to abrupt jumps, and the overall path lacks coherent temporal organization. This method violates both desirable properties: within-interaction frames do not form compact clusters, and cross-interaction transitions are neither smooth nor controlled. The behavior is consistent with severe inter-event drift, where the model is repeatedly pulled toward different condition-consistent modes without maintaining a stable connection to the prior trajectory.

\textbf{\textcolor{red}{\ding{184}} Mode-seeking persistence.}
The trajectory continues moving along a similar pattern regardless of whether frames belong to the current interaction or a new one, making within-event and cross-event transitions largely indistinguishable. Against our criteria, this method fails primarily on within-interaction aggregation: instead of forming compact local clusters for each prompt interval, the trajectory is dominated by a persistent motion tendency---particularly camera motion---that overrides event-conditioned structure. The model follows an internally favored motion pattern rather than reorganizing around each new event, constituting a form of mode-seeking dynamics.

\subsection{Further Latent Trajectory Comparisons}
\label{apx:vis-latent-traj}

\begin{figure*}[t]
    \centering
    \includegraphics[width=\textwidth]{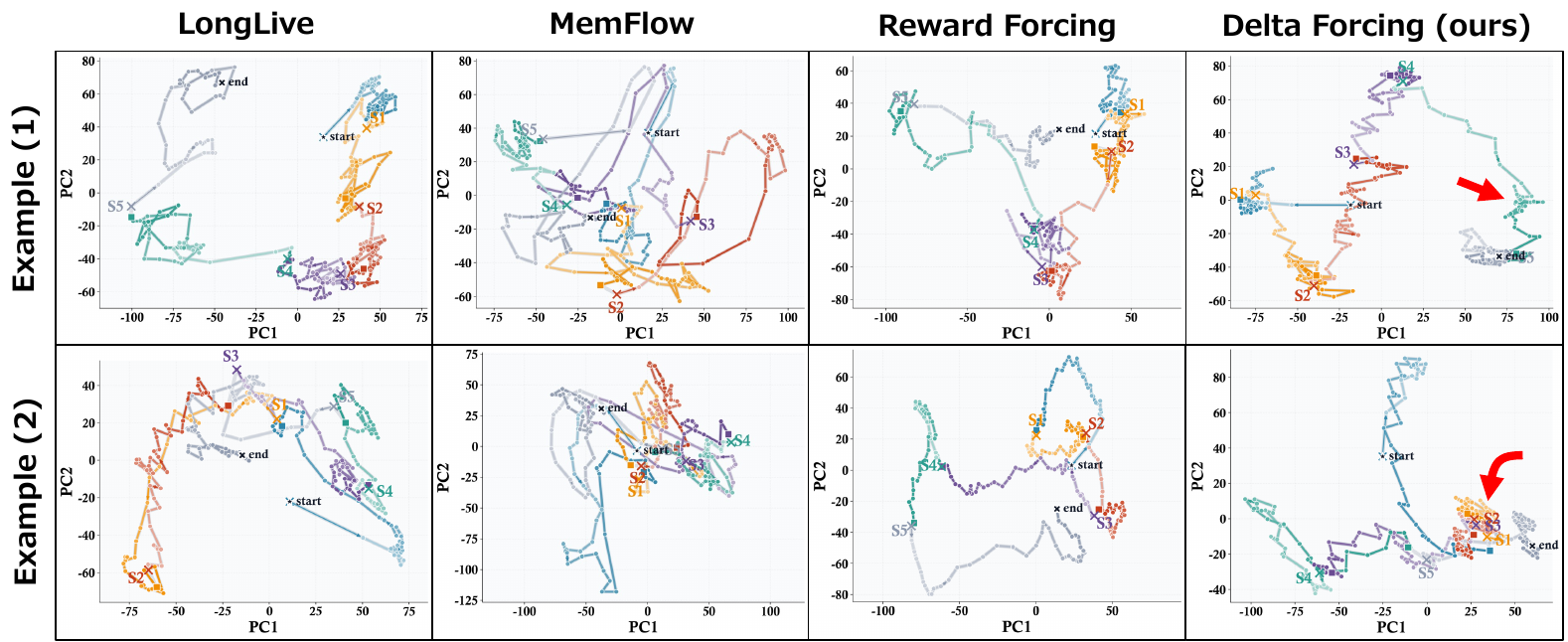}
    \caption{\textbf{Extended latent trajectory comparison.} Each row shows one example under the same multi-event prompt schedule, comparing three baselines (columns 1--3) against \framework (column 4). Red arrows highlight segments where \framework exhibits compact within-interaction clusters connected by smooth cross-interaction transitions, consistent with the desirable properties established in Section~\ref{sec:desirable}.}
    \label{fig:trajectory}
\end{figure*}

To verify that the trajectory patterns identified in Section~\ref{sec:observations} are consistent beyond a single instance, we extend the PCA-based latent trajectory analysis to additional examples and include \framework for comparison. Figure~\ref{fig:trajectory} presents two representative cases (rows), each comparing three baselines against \framework (rightmost column) under the same multi-event prompt schedule.

Across both examples, the baseline methods reproduce the failure modes previously identified. These patterns are consistent with the under-reactive, unstructured-drift, and mode-seeking behaviors discussed in Section~\ref{sec:observations}, confirming that these failure modes are systematic rather than instance-specific.

In contrast, \framework (rightmost column, red arrows) produces trajectories that more closely satisfy the desirable properties defined in Section~\ref{sec:desirable}. Within each interaction segment, the latent states remain relatively aggregated, forming compact local clusters. Across prompt switches, the trajectory transitions smoothly to a new region with sufficient displacement to reflect the semantic change. This combination of intra-state stability and responsive inter-state transitions provides latent-space evidence that \framework better balances within-interaction coherence and between-interaction adaptability.

\section{Experiment Details}
\label{apx:exp-details}

\noindent\textbf{Benchmark Description.}
This benchmark evaluates a video generation model's ability to produce long-form, narratively coherent video by decomposing a one-minute scene into six consecutive 10-second segments. Unlike single-shot text-to-video benchmarks that evaluate isolated clips, this benchmark stresses the model's capacity for cross-event consistency: maintaining stable character identity, environmental continuity, and a progressive storyline as the timeline unfolds.

\noindent\textbf{Formulation.}
Each test case is a sequence of prompts $\{P_1, P_2, \dots, P_N\}$ where each prompt $P_i$ corresponds to a fixed-length window $[t_i, t_i + \Delta t)$ with $\Delta t = 10$ seconds. The full benchmark sample spans $N \cdot \Delta t = 60$ seconds with $N = 6$ segments. Adjacent prompts $P_i$ and $P_{i+1}$ describe \textbf{consecutive events} in the same continuous scene --- there are no cuts, scene changes, or temporal jumps between segments. Persistent entities (characters, setting, key objects) are explicitly re-described in every prompt to anchor visual identity, while the \textbf{action and state evolve incrementally}.

\begin{tcolorbox}[calloutbox, title=Example@1 --- Young Wizard \& Dragon]
 
\timeline{00:00 -- 00:10}: A young wizard with a pointed hat and a sparkling robe is practicing magic in an ancient, stone-walled chamber. He waves a wooden wand, causing small objects on a table to float in the air. The chamber is lit by flickering torchlight, casting long shadows.
 
\vspace{6pt}
\timeline{00:10 -- 00:20}: A young wizard with a pointed hat in an ancient, stone-walled chamber continues to levitate objects with his wand. A small, scaly dragon with shimmering green wings flies into the chamber through a high window and lands on his shoulder. The wizard looks surprised but pleased.
 
\vspace{6pt}
\timeline{00:20 -- 00:30}: A young wizard with a pointed hat and a small green dragon on his shoulder practice magic together in the ancient, stone-walled chamber. He points his wand at a dusty book, and the dragon lets out a tiny puff of smoke, causing the book to flip open. The torchlight flickers on the walls.
 
\vspace{6pt}
\timeline{00:30 -- 00:40}: A young wizard with a pointed hat and a small green dragon on his shoulder in the ancient, stone-walled chamber laugh as the dragon sneezes, sending a shower of harmless sparks onto the floating objects. The wizard pats the dragon's head affectionately.
 
\vspace{6pt}
\timeline{00:40 -- 00:50}: A young wizard with a pointed hat and a small green dragon on his shoulder in the ancient, stone-walled chamber work together to create a larger spell. The wizard chants an incantation while the dragon flaps its wings, and a ball of swirling light forms between them.
 
\vspace{6pt}
\timeline{00:50 -- 01:00}: A young wizard with a pointed hat and a small green dragon on his shoulder in the ancient, stone-walled chamber watch as the ball of light floats up to the ceiling and bursts into a shower of magical stars. They look at each other and share a triumphant smile.
 
\end{tcolorbox}
 
\vspace{1em}
 
\begin{tcolorbox}[calloutbox, title=Example@2 --- Park Ranger \& Fawn, colback=green!5!white, colframe=green!50!black]
 
\timeline{00:00 -- 00:10}: A park ranger in a green uniform is walking along a dirt trail in a vast, sun-drenched national park. The landscape is filled with towering redwood trees and ferns. The air is clear and crisp.
 
\vspace{6pt}
\timeline{00:10 -- 00:20}: The park ranger walking in the vast national park discovers a small, orphaned fawn huddled at the base of a giant redwood tree. The fawn looks weak and scared. The forest is quiet.
 
\vspace{6pt}
\timeline{00:20 -- 00:30}: The park ranger and the small, orphaned fawn are in the vast national park. The ranger kneels down slowly, speaking in a soft, calm voice to avoid frightening the animal further. The redwood trees tower over them.
 
\vspace{6pt}
\timeline{00:30 -- 00:40}: The park ranger in the vast national park gently scoops up the small, orphaned fawn into their arms. The fawn is light and fragile. The sun filters through the massive redwood canopy.
 
\vspace{6pt}
\timeline{00:40 -- 00:50}: The park ranger in the vast national park carries the small, orphaned fawn carefully, cradling it against their uniform as they begin the long walk back to the ranger station. The forest path is dappled with sunlight.
 
\vspace{6pt}
\timeline{00:50 -- 01:00}: The park ranger walks with purpose through the vast national park, a protector of the forest and its creatures, carrying the small, orphaned fawn to safety. The giant redwood trees stand as silent witnesses.
 
\end{tcolorbox}

\section{User Study}
\label{apx:userstudy}

\begin{figure*}
    \centering
    \includegraphics[width=0.9\linewidth]{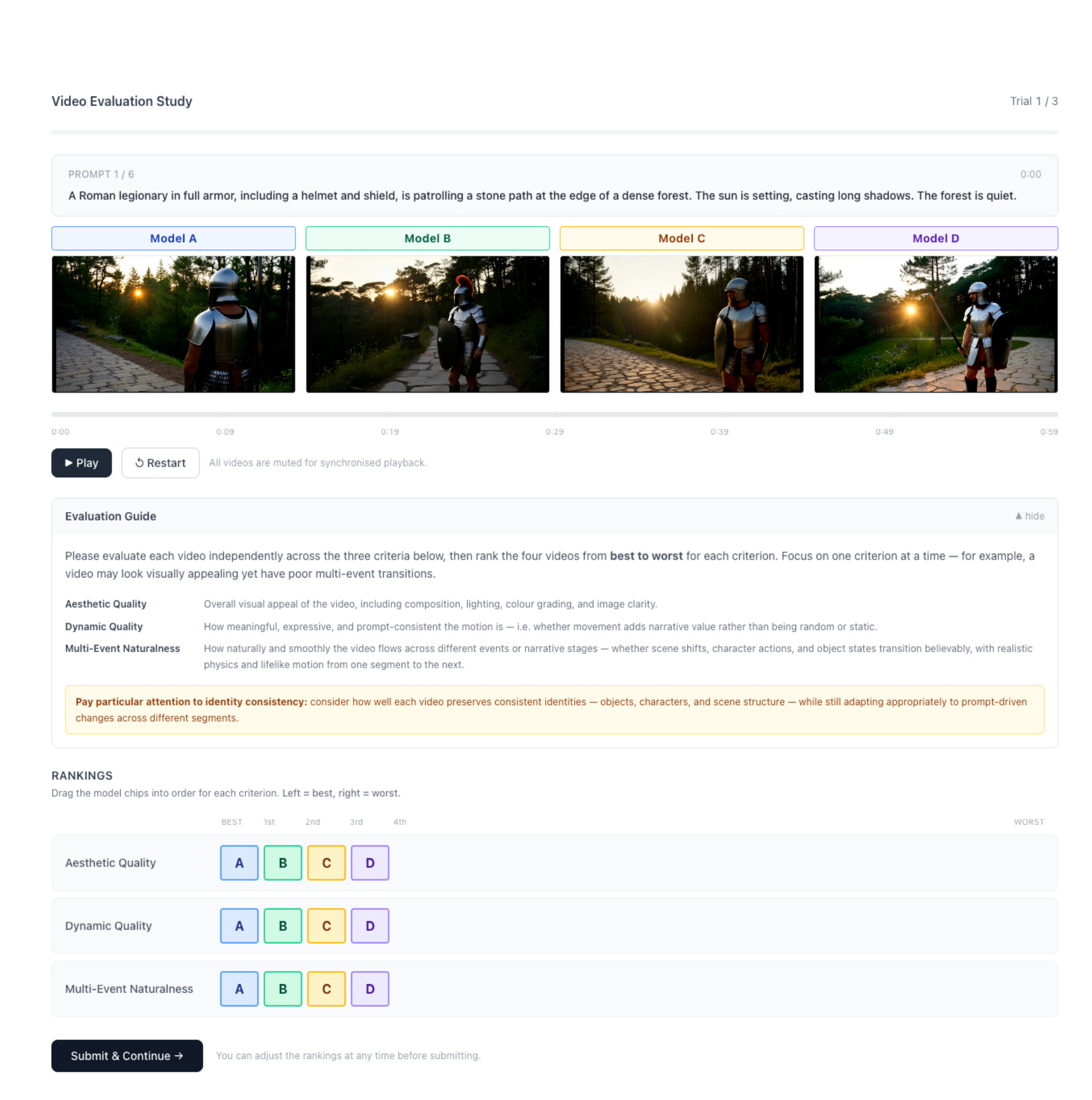}
    \caption{User study interface.}
    \label{fig:userstudy-interface}
\end{figure*}

Figure~\ref{fig:userstudy-interface} shows the web-based interface used for our human evaluation. For each trial, participants are shown a single text prompt at the top, followed by four videos generated by the compared models (anonymized as Model~A--D, with order randomized per trial). The videos are displayed side-by-side with synchronized playback controlled by shared \textit{Play} and \textit{Restart} buttons, allowing direct visual comparison. Below the videos, an evaluation guide briefly defines the three criteria: \textbf{Aesthetic Quality} (visual appeal, composition, lighting, and clarity), \textbf{Dynamic Quality} (meaningfulness and prompt-consistency of motion), and \textbf{Multi-Event Naturalness} (smoothness of transitions across narrative stages while preserving identity consistency). Participants are instructed to evaluate each criterion independently rather than producing a single overall judgment. For each criterion, they rank the four models from best to worst by dragging labeled chips into an ordered slot row, and submit rankings per trial.

